\documentclass[letterpaper,10pt,conference,final]{IEEEtran}

\usepackage[numbers]{natbib}
\usepackage{graphicx}       
\usepackage{algorithm}
\floatstyle{plaintop} \restylefloat{algorithm} 
\usepackage{algorithmic}
\usepackage{hyperref}
\usepackage{amsmath}
\usepackage{booktabs}
\usepackage{makecell}
\usepackage{comment}
\usepackage{arydshln}  
\usepackage[absolute]{textpos}
\usepackage{balance}

\usepackage{newtxtext}   
\usepackage{inconsolata} 

\hypersetup{colorlinks,linkcolor=blue,citecolor=blue,urlcolor=blue,bookmarks=true}
\hypersetup{pdftitle={A Spatio-Temporal Neural Network Forecasting Approach for Emulation of Firefront Models}}
\hypersetup{pdfauthor={Andrew Bolt, Carolyn Huston, Petra Kuhnert, Joel Janek Dabrowski, James Hilton, Conrad Sanderson}}

\renewcommand{\footnoterule}{%
  \kern -3pt
  \hrule width \columnwidth height 0.5pt
  \kern 2pt
}

\begin{document}

\title{A Spatio-Temporal Neural Network Forecasting Approach for Emulation of Firefront Models}

\author
  {%
  Andrew~Bolt\textsuperscript{{\tiny~}$\dagger$}, Carolyn~Huston\textsuperscript{{\tiny~}$\dagger$}, Petra~Kuhnert\textsuperscript{{\tiny~}$\dagger\diamond$}, Joel~Janek~Dabrowski\textsuperscript{{\tiny~}$\dagger$}, James~Hilton\textsuperscript{{\tiny~}$\dagger$}, Conrad~Sanderson\textsuperscript{{\tiny~}$\dagger\ddagger$}\\
  ~\\
  \textsuperscript{$\dagger$}{\tiny~}\textit{Data61 / CSIRO, Australia;}~
  \textsuperscript{$\diamond$}{\tiny~}\textit{Australian National University, Australia;}~
  \textsuperscript{$\ddagger$}{\tiny~}\textit{Griffith University, Australia}
  }

\maketitle{}

\begin{abstract}

Computational simulations of wildfire spread typically employ empirical rate-of-spread calculations
under various conditions (such as terrain, fuel type, weather).
Small perturbations in conditions can often lead to significant changes in fire spread
(such as speed and direction),
necessitating a computationally expensive large set of simulations to quantify uncertainty.
Model emulation seeks alternative representations of physical models using machine learning,
aiming to provide more efficient and/or simplified surrogate models.
We propose a dedicated spatio-temporal neural network based framework for model emulation,
able to capture the complex behaviour of fire spread models.
The proposed approach can approximate forecasts at fine spatial and temporal resolutions
that are often challenging for neural network based approaches.
Furthermore, the proposed approach is robust even with small training sets,
due to novel data augmentation methods.
Empirical experiments show good agreement between simulated and emulated firefronts, 
with an average Jaccard score of 0.76.

\end{abstract}


\begin{textblock}{13.44}(1.28,14.90)
\hrule
\vspace{1ex}
\noindent
\scalebox{0.725}{\textbf{{$^\ast$}~Published in:} International Conference on Signal Processing Algorithms, Architectures, Arrangements, and Applications, 2022. DOI:~\href{https://doi.org/10.23919/SPA53010.2022.9927888}{\tt 10.23919/SPA53010.2022.9927888}}
\end{textblock}

\IEEEpeerreviewmaketitle

\section{Introduction}

\begin{figure*}[t]
    \centering
    \includegraphics[width=0.9\textwidth]{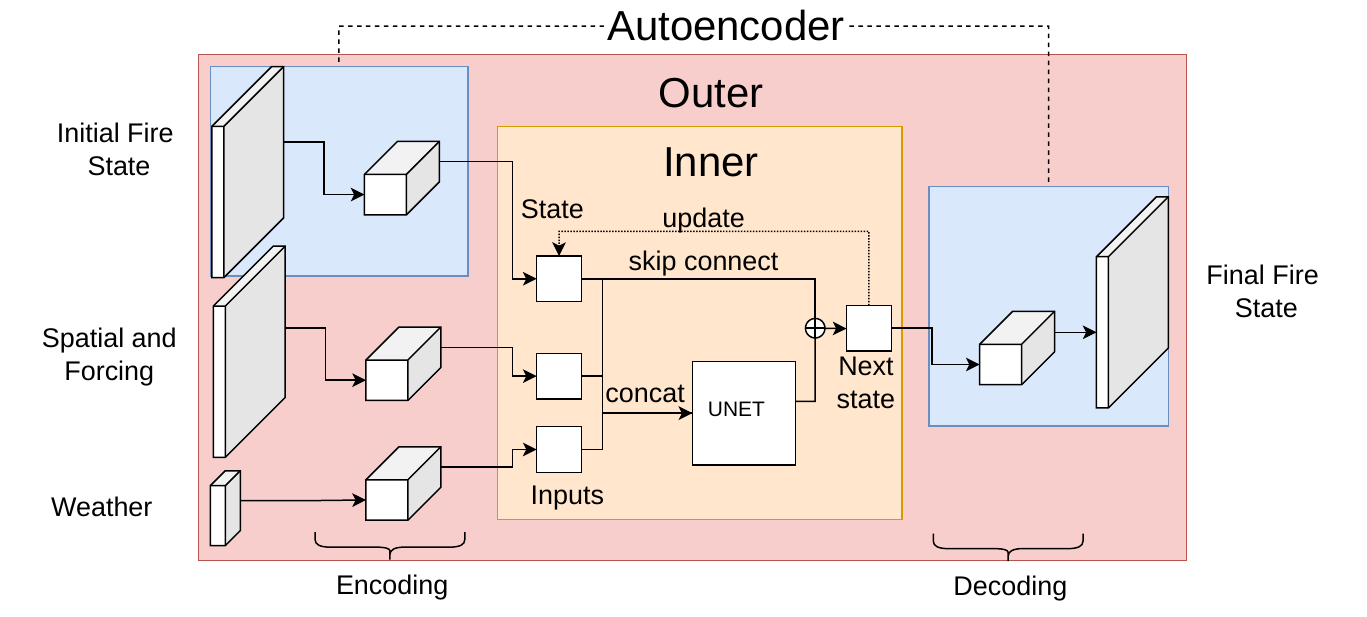}
    \caption{The proposed emulator architecture, comprising three main components. The autoencoder (blue) encodes and decodes the fire input state. The outer component (red) incorporates the autoencoder, as well as encoding spatial, forcing, and weather features. The inner component (orange) handles the dynamics of the emulator. The latent fire state is updated using information from the spatial and forcing layers, as well as weather feature inputs. These layers are concatenated and passed through a shallow U-Net structure. The sum of the U-Net output and the input latent fire state produce a new latent fire state estimate.
    }
    \label{fig:arch_overview}
\end{figure*}  

Wildfires pose a serious threat to communities as well as natural flora and fauna in many regions throughout the world~\cite{bot2022systematic}, \cite{Bowman2020VegetationFI},  \cite{scott2013fire}.
Forecasting spread of wildfires is critical in fire management, planning, and response efforts.
Simulated (\textit{in silico}) fires provide valuable data for operational managers to assess potential impacts on populated or sensitive areas, in order to guide active management, mitigation and evacuation efforts.

Several fire behaviour characteristic models have been developed \cite{finney1998farsite}, \cite{Spark2015}, \cite{Phoenix2008}.
Such models are generally computationally expensive as they may be based on complex methodologies such as the level-set method~\cite{Spark2015}. This can hinder their applicability for decision support, especially when large scale simulations or numerous ensemble predictions are required to account for uncertainty.

Model emulation (also known as surrogate modelling) employs a computationally efficient predictive model that approximates a complex physical process model, such as computer or numerical simulators~\cite{alizadeh2020managing}, \cite{sanderson2021emulation}. Emulation may be able to overcome some of the limitations of large scale complex simulations. Early emulation approaches used machine learning techniques such as Gaussian processes~\cite{kennedy2001},
followed by random forests~\cite{gladish2018}, \cite{leeds2013}
and deep neural networks~\cite{Kashinath2021}, \cite{kasim2021}, \cite{sit2020}. Neural networks are highly adaptable and have been successfully implemented in several physical system emulation problems~\cite{kasim2021}, \cite{sit2020}, \cite{thiagarajan2020designing}.

Recent reviews on applications of machine learning to wildfires cover fire susceptibility prediction, fire spread prediction, fuel categorisation, fire occurrence detection, and decision support systems~\cite{abid2021survey}, \cite{bot2022systematic}, \cite{jain2020review}.
Deep learning architectures such as convolutional neural networks (CNNs) \cite{allaire2021}, \cite{hodges2019wildland}, \cite{radke2019firecast}, and recurrent neural networks~\cite{burge2020convolutional} have also been applied. 


Within the literature on neural networks related to model emulation for fire spread and growth prediction, 
Allaire et al.~\cite{allaire2021} present a CNN emulator for hazard assessment in a contained region of interest. The emulator predicts the amount of burned land (scalar value). The model does not estimate fire dynamics. 
Burge et al.~\cite{burge2020convolutional} and Hodges et al.~\cite{hodges2019wildland} propose CNN emulators for predicting fire dynamics. Both approaches use a small output array size ($<$~100 pixels) which limits the spatial resolution or extent that can be evaluated. 
Radke et al.~\cite{radke2019firecast} propose a CNN based emulator, which estimates the likelihood of a pixel outside the firefront burning within a 24 hour window. Rather than evaluating each pixel, the likelihoods of a set of sample pixels are generated. The wide temporal resolution limits the ability of the model to estimate fine timescale dynamics of the fire.
Sung et al.~\cite{sung2021} propose a neural network model incorporating a U-Net structure~\cite{ronneberger2015},
trained with a dataset of daily fire perimeters.
The model estimates the likelihood of the fire reaching a given pixel.
Relatively low accuracy is obtained, possibly due to low temporal resolution.

In this paper we propose a neural network based emulator for estimating high resolution fire spread over large spatial and temporal domains.
The model is trained with simulated data produced using empirical rate-of-spread estimates.
The proposed emulator design is able to incorporate data of varying spatial and temporal resolutions. Furthermore it is capable of generating estimates over large spatial extents with varying shapes, which is often challenging for emulation approaches.
Lastly, the model is robust on small training sets due to employing novel data augmentation methods.

The architecture of the proposed emulator is overviewed in Section~\ref{sec:modelling},
and its design features are covered in Section~\ref{sec:design}.
As~there are hyper-parameters that can be adjusted,
we provide an empirical evaluation at various configurations in Section~\ref{sec:evaluation}.
The best performing configuration has an average Jaccard score of 0.76,
indicating good agreement between simulated and emulated firefronts.
The proposed model provides a template upon which further developments can be introduced,
especially methods for uncertainty quantification.

\section{Emulator Architecture}
\label{sec:modelling}

Fig.~\ref{fig:arch_overview} shows a simplified schematic of the proposed emulator architecture. There are three main components: autoencoder component, outer component, and inner component.
The autoencoder component is trained separately and its weights are transferred to parts of the outer component. The outer component is a feature engineering and downsampling network which passes inputs to the inner component. The outer component also upsamples the outputs from the inner component. Finally, the inner component handles the dynamics of the system. It forecasts latent fire states based on information from spatial and temporal features. 

There are three types of inputs to the model: 
fire state image, spatial data and forcing terms, and finally weather time series.
Each fire state is an image where pixel values represent when the fire reached a given location.
The spatial data are images representing heightmaps and land classes (eg.~forest, grassland).
The forcing terms are curing and drought factors.
The weather time series include temperature, wind speed, and relative humidity.

\subsection{Outer Network}
\label{sec:outer_network}

The outer components of the model are illustrated by the red shading in Fig.~\ref{fig:arch_overview}. This component incorporates the fire state autoencoder. Convolutional layers are used on the spatial data (heightmap and land class map) for feature extraction; downscaling is performed by strided convolutional layers. This encodes the information into a smaller latent representation. Similarly, forcing inputs (drought and curing factors) are expanded to have the same spatial extent as the latent spatial features and are concatenated together.

The downsampling and upsampling of the fire state is performed by the autoencoder. The latent fire state and  latent spatial features form the state component of the inner network.

Weather features are in the form of time-series data. Interpolation via upsampling is performed. Each set of weather values at a given time are passed through two dense layers for feature extraction before being expanded to the same spatial extent as the latent spatial features. These are then passed to the inner network as inputs.

\subsection{Autoencoder}
\label{sec:architecture}

The purpose of the autoencoder is to encode and decode the fire state to and from a lower latent dimension. The autoencoder is represented by joining the blue components in Fig.~\ref{fig:arch_overview}. The encoding component consists of only linear transformations to preserve the relative temporal relationships between pixel values. An average pooling layer is used before a space to depth transformation. The decoding component consists of a depth to space transformation followed by a strided convolutional layers. The autoencoder is trained on fire state data separately from the full emulator.

\subsection{Inner Network}
\label{sec:inner_network}

The inner component of the model incorporates the dynamics of the system, taking an initial latent fire state and producing a new updated estimate. The inner component is illustrated by the orange shading in Fig.~\ref{fig:arch_overview}. This component is a one-step ahead forecast module. 

The state of the module consists of the latent fire arrival state. This is updated using dynamic inputs from the latent weather features and static inputs from the latent spatial and forcing features. Each successive weather input advances the fire state forward in time.

The inputs and fire state are concatenated and passed into a shallow U-Net~\cite{ronneberger2015}, \cite{wang2019}. U-Nets are formed by joining a contracting path with an expanding path, formed by down/upsampling convolutional layers. This structure is able to capture dynamics at various spatial resolutions, with deeper levels capturing broader interactions. We incorporate a skip connection between the input latent fire state and the output of the U-Net. In effect the U-Net only needs to learn the change or \textit{residual} between successive latent fire states.

\section{Design Features}
\label{sec:design}

In this section we will briefly discuss some of the unique challenges that inform our choice of emulator design. Broadly, we want an emulator that is agnostic to temporal and spatial resolutions and is able to generate high resolution outputs.

In order to achieve a model that can incorporate various spatial and temporal resolutions we express features like distance, height, and wind speed in unit-less terms. For example, wind speed is converted from meters per second into pixel lengths per interval. In this way a different dataset operating with a different spatial (or temporal) resolution can be re-sampled to be compatible with the model inputs. 

A further challenge is that the training data often consists of large sized image arrays ($>$~1000~px). Furthermore, the image arrays can have varying sizes. To address this we choose a fully convolutional network, and encode the spatial data into a smaller latent representation through strided convolutional operations. This reduces the complexity of the representation and allows the model to take arbitrarily sized inputs. This is performed by the outer component of the model (Section~\ref{sec:outer_network}). In addition, the inner component (Section~\ref{sec:inner_network}) employs a shallow U-Net structure. This allows for long range interactions between pixels to be considered by the model using only a modest overhead of complexity.

The inner component of the emulator deals with the dynamic changes to the fire arrival state. To ensure that the outer model does not produce any dynamic changes, we use an autoencoder. The autoencoder is trained on the fire arrival state. Once trained, the autoencoders' weights are frozen. The autoencoder is then used by the outer network to encode and decode the fire arrival state.

Wildfire spread is an inherently stochastic process. In some cases an estimate may incorrectly place the fire slightly ahead of an obstacle such as a body of water. If we consider this as the source of a new firefront propagating through open terrain, then the burned area by the new front will increase quadratically in time~\cite{Zekri2016}. By training the model over a single time interval we limit the training penalty of these mismatches.

As an additional step to the training process, each sample is cropped around a point on the fire’s perimeter. This reduces the spatial extent of each sample and generates a uniform size. This allows for batch processing as well as greatly reducing the memory requirements for training. 

Using cropping and single intervals for model training represents a novel method of data augmentation. A single fire can produce numerous semi-independent training samples by using various cropping locations and time intervals. Further data augmentation was performed using rotation, reflection and transposition transformations.

A drawback the cropping approach is that cropping removes some information about the fire's position as a whole. The likelihood of fire `entering’ a cropped region cannot be inferred during training. To account for this we pad each sample. In this padded region, each pixel value is assigned to the maximum of the predicted or target values. This removes the loss penalty when the presence of fire is not correctly inferred around the border of a region.

\section{Empirical Evaluation}
\label{sec:evaluation}

In this section we first overview the evaluation dataset,
and then present an empirical evaluation using several configurations of hyper-parameters in the proposed emulator.

\subsection{Dataset}
\label{subsec:dataset}

\begin{figure*}
\centering
\begin{minipage}{0.49\textwidth}
  \centering
  \includegraphics[width=\textwidth]{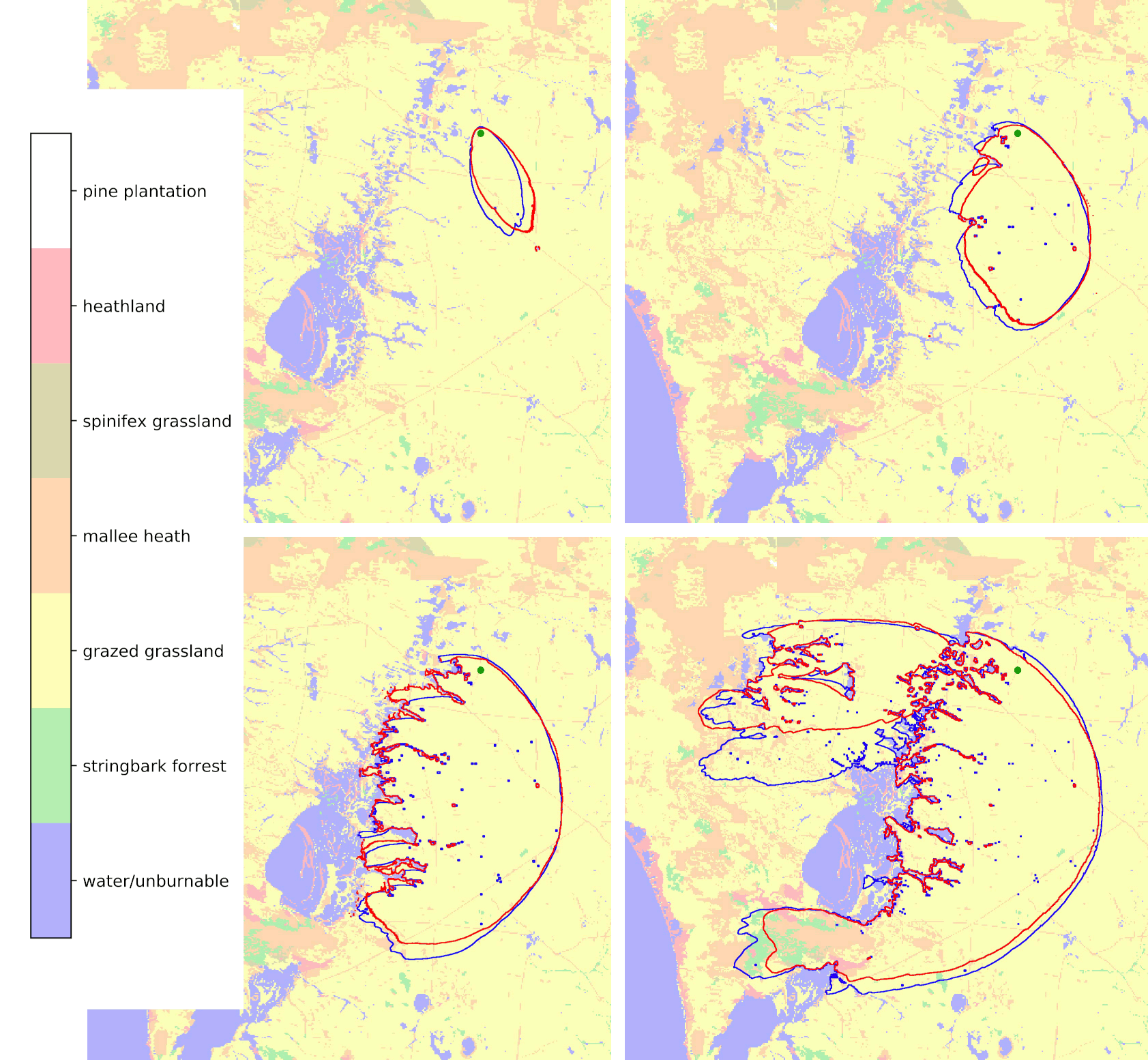}
  \caption{Evolution of firefront contours for a trial, shown over four panels (left-to-right, top-to-bottom). Emulator (red), simulation (blue) and ignition point (green) are overlaid over land classes. Dominant land classes are grassland (yellow), mallee-heath shrubland (orange), and water (blue). The wind initially drives the fire south, before turning west. Map size is 46.1~km~$\times$~46.1~km, 30 meter resolution.}
  \label{fig:quad}
\end{minipage}%
\hfill
\begin{minipage}{0.49\textwidth}
  \centering
  \includegraphics[width=\textwidth]{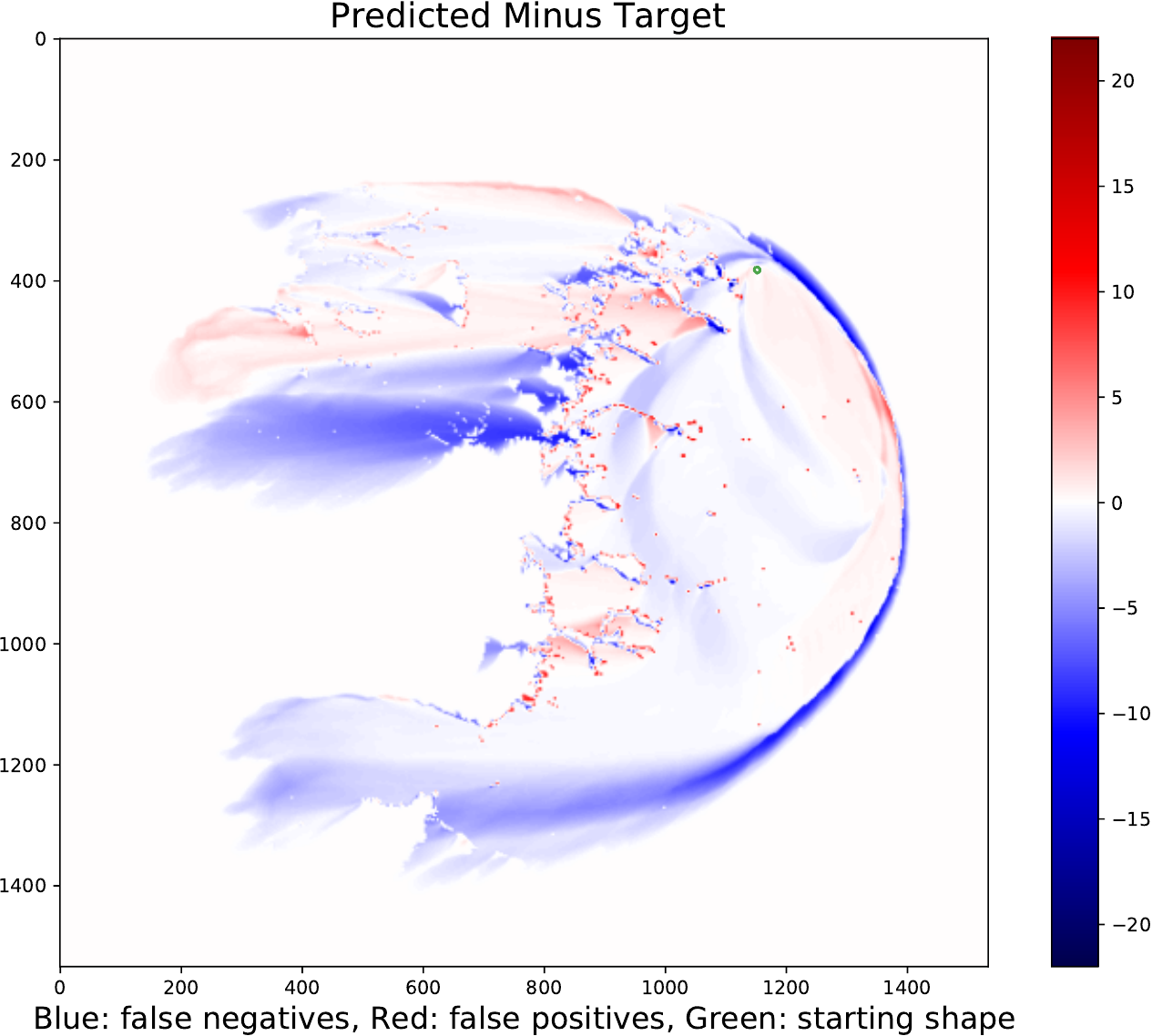}
  \caption{The difference between predicted and target fire arrival times (measured in 30 minute intervals) for the same test sample as illustrated in Fig.~\ref{fig:quad}. Positive values (red) indicate false-positives while negative values (blue) represent false-negatives. The Jaccard score for this trial is 0.81.}
  \label{fig:difference}
\end{minipage}
\end{figure*}

\begin{figure*}
\centering
\begin{minipage}{0.49\textwidth}
  \centering
  \includegraphics[width=\textwidth]{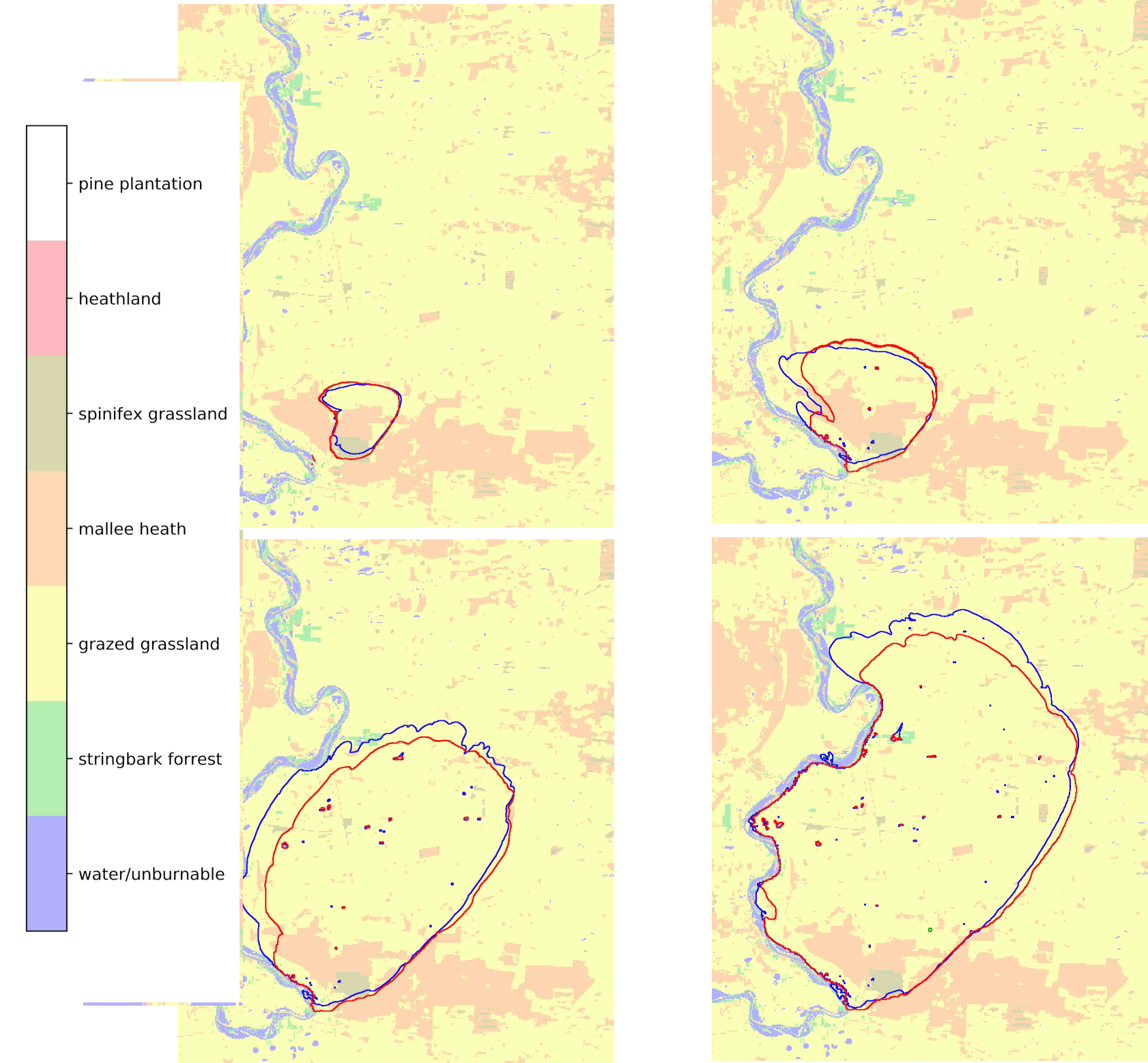}
  \caption{Evolution of firefront contours for a trial, shown over four panels (left-to-right, top-to-bottom). Emulator (red), simulation (blue) and ignition point (green) are overlaid over land classes. Dominant land classes are grassland (yellow), mallee-heath shrubland (orange), and water (blue). The wind initially drives the fire south-east, before turning north. Map size is 46.1~km~$\times$~38.4~km, 30 meter resolution.} 
  \label{fig:quad_b}
\end{minipage}%
\hfill
\begin{minipage}{0.49\textwidth}
  \centering
  \includegraphics[width=\textwidth]{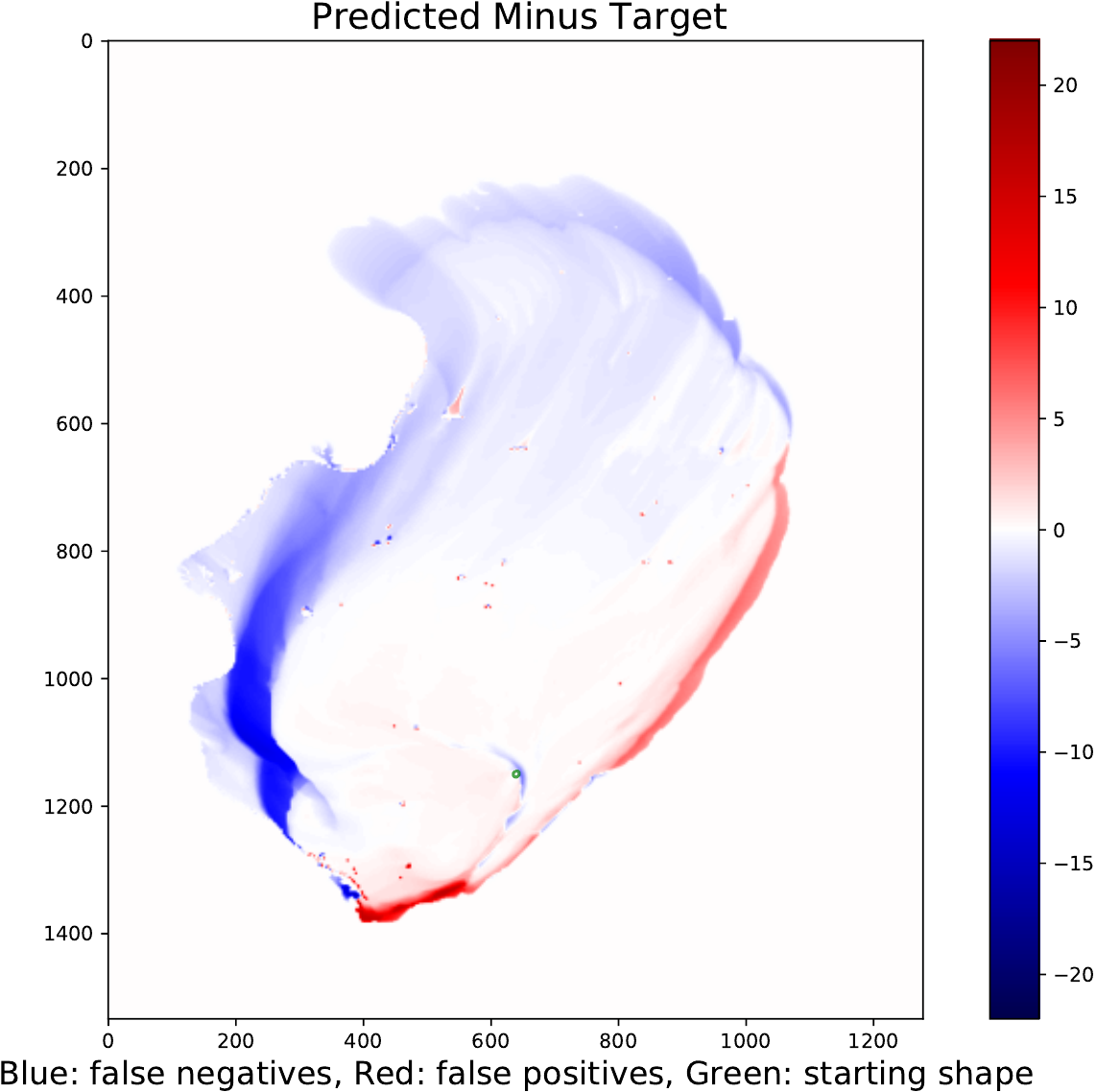}
  \caption{The difference between predicted and target fire arrival times (measured in 30 minute intervals) for the same test sample as illustrated in Fig.~\ref{fig:quad_b}. Positive values (red) indicate false-positives while negative values (blue) represent false-negatives. The Jaccard score for this trial is 0.90.}
  \label{fig:difference_b}
\end{minipage}
\end{figure*}

We use a dataset of 195 simulated fires generated via the SPARK fire simulation platform~\cite{Spark2015}.
155 fires are used as the training set; 40 fires are reserved for validation.
During training and validation the regions are cropped and only a single interval is used.
Furthermore, we use a prediction set, composed of the 40 validation fires, but no cropping is performed and the entire duration of the fire is considered. This set is not used in training, but is used to evaluate the performance of the emulator across the full scope of a fire scenario.

The location of the fires is representative of regional South Australia. Land classification maps\footnote[1]{Land classification datasets  derived from Department of Agriculture and Water Resources (ABARES) Land Use of Australia 2010-11 dataset. Data is publicly available under Creative Commons Attribution 3.0 Australia Licence.}, topology data\footnote[2]{Topography data sets derived from Geoscience Australia SRTM-derived 1~Second Digital Elevation Models Version 1.0. Data is publicly available under Creative Commons Attribution 4.0 International Licence.}, and weather conditions\footnote[3]{Meteorological time series data sets derived from Bureau of Meteorology automated weather station data.} are sampled to reflect realistic regional conditions. There is a large bias in land classification towards grassland (78.8\%), mallee-heath shrubland (10.5\%), water/unburnable (6.27\%). The weather is representative of high fire risk conditions, often consisting of high temperatures, low humidity, and moderate to high winds. 

Spatial data is converted into the same coordinate reference system and has a resolution of 30 meters. The height map is converted into $x$ and $y$ gradient maps. Weather data is polled from weather stations every 30 minutes (one interval). The weather values are interpolated (upscaled) into 4 slices for the majority of our trials. Values are scaled by maximum and minimum values in the training set.
Wind speed and direction are converted into $x$ and $y$ components. Finally there are two forcing terms that the model incorporates: \textit{drought factor} and \textit{curing factor}. These terms depend on long term weather trends that are considered fixed for the duration of the fire.

\subsection{Metrics and Loss}
\label{subsec:metrics_and_loss}

For evaluation metrics we choose the Jaccard score, also known as intersection over union (IOU) score. 
While this straightforward metric may not capture the whole `goodness of fit', it nevertheless provides a basic grounding~\cite{huston2015fit}. 

The autoencoder component is trained using mean absolute error (MAE) loss. For training the emulator we introduce a custom loss function, which evaluates how well the predictions perform against a benchmark trivial prediction (where the output is the same as the input). Let $y_i$, $y_t$, and $y_p$ be the initial fire state, target fire state, and predicted fire state respectively. Furthermore, let MAE$(a, b)$ be the mean absolute error across corresponding pixels in images $a$ and $b$. The loss $\mathcal{L}$ of fire state~$\mathcal{P}$ is defined as:

\noindent
\begin{align}
    \mathcal{L} (\mathcal{P}) = \mathrm{log}_{10}\left(\frac{\mathrm{MAE}(y_p, y_t) + \tau}{\mathrm{MAE}(y_i, y_t) + \tau} \right).
    \label{eq:loss}
\end{align}

\noindent
Small fire growth results in only a small set of pixels indicating burns,
which in turn leads to very small MAE$(y_p, y_t)$.
To address this, MAE$(y_i, y_t)$ is used as a normalisation factor.
Furthermore, $\tau = 10^{-12}$ is used to remove singularities that arise if either MAE values approach zero.

\subsection{Evaluation}
\label{subsec:evaluation}

We implemented the emulator in TensorFlow~\cite{tensorflow_usenix_2016}, using the  \textit{Adam} optimiser~\citep{Diederik2015} with a batch size of~16. The autoencoder was trained for 20 epochs and returned an MAE loss of $2.1\times 10^{-3}$. The emulator was trained for 50 epochs using the loss function in Eqn.~(\ref{eq:loss}).

We train the emulator under several configurations of hyper-parameters.
Specifically, we test variously sized cropping windows (\textit{c}),
U-Net depths (\textit{d}), and padding values (\textit{p}).
The number of interpolation slices per interval is 4.

Table~\ref{tab:evaluation} shows the average loss and Jaccard (IOU) scores for the training and validation sets, as well as for the prediction set. The prediction results are split into two parts. The first part is when the emulator is run over the entire duration of the prediction set (intervals 0-22). The second part is when the emulator begins with the fire in progress (intervals 5-22). Furthermore, two Jaccard scores are shown: the top value represents an unweighted score averaged over all samples, while the lower bracketed value shows the score weighted by burned area over all samples.

\begin{table}
    \caption
      {
      Model loss and evaluation metrics.
      Each configuration comprises \textit{c}, \textit{d}, \textit{p} components, where \textit{c} is size of the cropping window, \textit{d} is the depth of the U-Net component, and \textit{p} is the amount of padded pixels.
      }
    \setlength{\tabcolsep}{5pt}
    \label{tab:evaluation}
    \centering
    \begin{tabular}{lcccccc}
            \toprule
            \makecell{Configuration}& \makecell{Train. \\ loss}     & \makecell{Val. \\ loss}          & \makecell{Train. \\ Jacc.} &  \makecell{Val.\\ Jacc.}  & \makecell{Pred. Jacc.\\ (0-22)} & \makecell{Pred. Jacc.\\ (5-22)}\\
            \midrule
            \makecell{512c, 1d, 32p}& -1.43& -1.43   &  0.78   & 0.79     & \makecell{\textbf{0.76} \\ \textbf{(0.80)}}& \makecell{\textbf{0.79} \\ \textbf{(0.83)}}    \\ \hdashline[0.5pt/5pt]
            \makecell{512c, 1d, 64p}& -1.47   & -1.48  & 0.79          & 0.79 & \makecell{0.71 \\ (0.74)} & \makecell{0.77 \\ (0.74)}      \\ \hdashline[0.5pt/5pt]
            \makecell{512c, 2d, 32p}&  -1.43   & -1.45  & 0.78          & 0.79 & \makecell{0.71 \\ (0.74)} & \makecell{0.77 \\ (0.81)}      \\ \hdashline[0.5pt/5pt]
            \makecell{512c, 2d, 64p}&  -1.55   & -1.51  & 0.81          & 0.80 & \makecell{0.69 \\ (0.70)} & \makecell{0.71 \\ (0.76)}      \\ \hline
            \makecell{256c, 1d, 32p}&  -1.61   & -1.73  & 0.81          & 0.83 & \makecell{0.64 \\ (0.67)} & \makecell{0.69 \\ (0.72)}      \\ \hdashline[0.5pt/5pt]
            \makecell{256c, 1d, 64p}&  -1.88   & -1.92  & 0.86          & 0.86 & \makecell{0.71 \\ (0.72)} & \makecell{0.72 \\ (0.77)}      \\ \hdashline[0.5pt/5pt]
            \makecell{256c, 2d, 32p}&  -1.64   & -1.67  & 0.82          & 0.82 & \makecell{0.67 \\ (0.74)} & \makecell{0.71 \\ (0.77)}      \\ \hdashline[0.5pt/5pt]
            \makecell{256c, 2d, 64p}&  -1.98   & -1.93  & 0.87          & 0.86 & \makecell{0.65 \\ (0.68)} & \makecell{0.68 \\ (0.74)}      \\ \hline
            \makecell{128c, 1d, 32p}&  -1.91   & -1.92  & 0.86          & 0.86 & \makecell{0.42 \\ (0.43)} & \makecell{0.45 \\ (0.45)}      \\ \hdashline[0.5pt/5pt]
            \makecell{128c, 2d, 32p}&  -1.77   & -1.82  & 0.84          & 0.85 & \makecell{0.52 \\ (0.52)} & \makecell{0.53 \\ (0.55)}      \\
            \bottomrule
    \end{tabular}
\end{table}

Direct comparison between configurations of loss values is not possible due to differences in how padding and cropping sizes affect the loss function defined in Eqn.~(\ref{eq:loss}). We note that the differences in loss and Jaccard scores between training and evaluation sets are small. This indicates that the model is generalising well to the entire dataset. 

The prediction set shows how well the model performs across the full spatial and temporal extents of each sample. The best performance is found for larger cropping sizes (512 pixels). The prediction Jaccard scores of the smaller cropped samples are significantly lower than that found during training and evaluation. There does not appear to be a significant difference between depth 1 and 2 U-Net configurations. Finally, the model performs better when using a 32 pixel padding distance. A number of trials were also performed using 6 interpolation slices per interval rather than 4; only a very marginal improvement in Jaccard scores was observed.

Figs.~\ref{fig:quad} and~\ref{fig:difference} illustrate the emulator performance on a sample fire. There is close agreement between simulated and emulated fires for much of the duration. The largest disagreement occurs as the fire passes through a region containing small bodies of water. The emulator fails to find a similar pathway to the original Spark simulation. 

Figs.~\ref{fig:quad_b} and~\ref{fig:difference_b} show the emulator performance on another sample fire. In this example there is also good agreement between simulated and emulated fires. To the west an early under-estimation by the emulator leads to a large under-estimation as the wind changes and pushes the fire north.

These types of behaviour are typical of many samples that have been manually inspected. Small differences between emulator and simulation are often exaggerated over time. Nonetheless, the overall dynamics of the emulator appear to be in line with expected fire behaviour.

\section{Conclusion}
\label{sec:conclusion}

In this paper we have shown how convolutional networks can be constructed in order to closely emulate wildfire spread from the Spark simulator,
resulting in an average Jaccard (IOU) score of 0.76 for up to 11.5 hour fire duration. Qualitatively, the emulator makes predictions that exhibit very similar behaviour to that of the targeted simulations. The stochastic nature of wildfires means that large discrepancies are often the result of small differences being exaggerated, rather than a fundamental problem in the emulation estimate. 

The proposed approach has several features that make it versatile. It is able to work using variable spatial extents and resolutions, variable temporal extents and resolutions, as well as being able to incorporate various types of spatial, temporal and scalar features. 

We use a novel approach to model training. This approach incorporates transfer learning as well as data augmentation in the form of targeted cropping and the use of single intervals for training (rather than full duration trials). Additionally, we use a custom loss function that is designed to operate well for this specific class of problem.

The flexibility of the modelling approach means that it should be possible to incorporate new features into the model without needing to fully retrain the model. For example, if a new land class is added, then it is possible that only the first few downsampling convolutional layers would need to be retrained.

A further area of interest is to take data from real world samples and use these to `fine tune' the model parameters. In this way it may be possible to use relatively sparse real world data to improve model performance, and potentially infer fire dynamics directly.

The proposed approach is not inherently specific to wildfire prediction. Similar geo-spatial modelling problems such as pollutant spread, pest spread, or disease spread may also be well represented by a similar emulation approach.

In followup work we have explored using ensemble simulations as training data for emulators, in order to directly generate likelihood estimates of fire locations~\cite{Bolt_2023}.

\small
\def~{\,}  
\bibliographystyle{ieee}
\bibliography{references}

\end{document}